\documentclass{article}
\usepackage{spconf,amsmath,graphicx}
\usepackage{multirow,stmaryrd,url,paralist}
\usepackage{pifont}% http://ctan.org/pkg/pifont
\newcommand{\cmark}{\ding{51}}%
\newcommand{\xmark}{\ding{55}}%
\usepackage[table]{xcolor}

\title{Abstractive Dialogue Summarization with \\Sentence-Gated Modeling Optimized by Dialogue Acts}
\name{Chih-Wen Goo$^{\dag\star}$ \thanks{This work is done while the author was at National Taiwan University.} and Yun-Nung Chen$^\dag$}
\address{$^\dag$National Taiwan University, Taipei, Taiwan\\
$^\star$MediaTek Inc., Hsinchu, Taiwan
\\ {\tt \normalsize r05944049@ntu.edu.tw\quad y.v.chen@ieee.org}}
\begin{document}
%\ninept
%
\maketitle
\begin{abstract}
Neural abstractive summarization has been increasingly studied, where the prior work mainly focused on summarizing single-speaker documents (news, scientific publications, etc).
In dialogues, there are diverse interactive patterns between speakers, which are usually defined as dialogue acts.
The interactive signals may provide informative cues for better summarizing dialogues. 
This paper proposes to explicitly leverage dialogue acts in a neural summarization model, where a sentence-gated mechanism is designed for modeling the relationships between dialogue acts and the summary.
The experiments show that our proposed model significantly improves the abstractive summarization performance compared to the state-of-the-art baselines on the AMI meeting corpus, demonstrating the usefulness of the interactive signal provided by dialogue acts.\footnote{The source code is available at \url{http://github.com/MiuLab/DialSum}.}
\end{abstract}
\section{Introduction}
\label{sec:intro}

With a large amount of textual information available, text summarization has been widely studied for several years in natural language processing, which can be categorized into two types: \emph{extractive summarization} and \emph{abstractive summarization}.
Extractive methods assemble the summary from the source text directly~\cite{Kupiec1995,chen2011spoken,saggion2012}, while abstractive methods generate words to form the summary~\cite{liu2015towards,chopra2016abstractive,NallapatiXZ16}.
With the rising trend of neural models, abstractive summarization has been widely investigated recently~\cite{rush2015neural,chopra2016abstractive}.
In addition, some recent work proposed to combine advantages from two types of methods and achieved better summarization results~\cite{gu2016,miao2016,see2017get,hsu2018unified}.

Most of the summarization work focused on single-speaker written documents such as news, scientific publications, etc~\cite{lee2014spoken,rush2015neural,gehrmann2018bottom}.
In addition to text summarization, speech summarization is equally important especially for spoken or even multimedia documents, which are more difficult to browse than text, such as multi-party meetings.
Therefore, speech summarization has been investigated in the past~\cite{maskey2005comparing,harwath2012topic,riedhammer2010long,chen2011automatic,chen2012two,chen2013multi}.
However, almost all prior work focused on summarizing the documents based on the mentioned salient content instead of the interactive status, but this behavioral signal should be important for dialogue summarization.

To better summarize a meeting, not only the content but also the inter-speaker interactions are important.
Prior dialogue summarization work utilized prosody or speaker information as interactive patterns for better extracting salient sentences~\cite{chen2012two,chen2012intra}.
However, abstractive summarization for dialogue/meeting summarization has not yet explored due to the lack of suitable benchmark data~\cite{gambhir2017recent}, because the benchmark dialogue data is only annotated with the importance of utterances without abstractive summaries~\cite{mccowan2005ami}.
In order to bridge the gap, this paper benchmarks the abstractive dialogue summarization task using  the AMI meeting corpus~\cite{mccowan2005ami}, where the summaries are produced based on the annotated topics the speakers discuss about.
A \emph{topic} or a \emph{high-level description} of a meeting is treated as the abstractive summary; for instance, ``\textit{evaluation of project new idea for TV}'' is a summary of the meeting topics.
Such dialogue summaries are very short and may not contain words directly mentioned by the speakers, making automatic summarization more challenging.

\begin{figure*}[t!]
\centering
\begin{tabular}{|l|c|}
\hline
\multicolumn{1}{|c|}{\bf Multi-Party Dialogue} & \bf Dialogue Act\\
\hline\hline
A: mm-hmm . & Backchannel \\
B: mm-hmm . & Backchannel \\
C: then , these are some of the remotes which are different in shape and colour , but they have many buttons . & Inform \\
C: so uh sometimes the user finds it very difficult to recognise which button is for what function and all that . & Inform \\
D: so you can design an interface which is very simple , and which is user-friendly . & Inform \\
D: even a kid can use that . & Inform \\
A: so can you got on t t uh to the next slide . & Suggest \\
\hline
\multicolumn{2}{|l|}{Summary: alternative interface options} \\
\hline
\end{tabular}
\caption{A dialogue instance in the dataset built from the AMI meeting corpus.}
\label{fig:example_data}
\end{figure*}

A dialogue is a sequence of utterances interacting between multiple participants, where each utterance would modify both participants' cognitive status and the current dialogue state.
The effect of an utterance on the context is often called a \emph{dialogue act}~\cite{Bunt94}, which provides informative cues for better understanding dialogues.
Therefore, dialogue act classification has been widely studied in the spoken language understanding research field, and previous work about dialogue act recognition used information sources from multiple modalities, including linguistic information, global contextual properties like knowledge about participants, and so on~\cite{Samuel1998,wright1999,Stolcke00,Kluwer2010,tran2017preserving}.
Popular approaches for dialogue act classification include support vector machine (SVM)~\cite{tavafi2013}, Naive Bayes~\cite{Stolcke00,Keizer2002,Ang05}, logistic regression~\cite{chen2013empirical}, and recurrent neural network (RNN)~\cite{ji16,Khanpour16,lee2016,Kalchbrenner13}.

Dialogue act classification and summarization are usually treated independently and used for different goals.
In this paper, we leverage dialogue act information to improve dialogue summarization.
\iffalse
Imagine that we have a meeting recording lasting for an hour.
One needs a sliding window to go through this very long conversation to predict the dialogue acts.
Since the conversation focus in these small windows may vary from serious discussion to off-topic chitchat, it makes the model harder to have a universal well solution without knowing the ``topic'' of windows.
From another point of view, a "topic" or a "high level description", for instance, \textit{evaluation of project new idea for TV} can be modeled as a summary of the utterances in one window.
Hence, we have shown that dialogue act recognition and summarization are related and exist the opportunity to be solved in one model.
\fi
Assuming that dialogue acts, indicating interactive signals, may be important for better summarization, how to effectively integrate the information into a neural summarization model is the main focus of this paper.
Prior work attempted at modeling the discourse information and proposed a discourse-aware summarization model using the hierarchical RNN~\cite{cohan2018discourse}, where the between-utterance cues are modeled in an implicit way.
Also, they performed the model in a publication summarization task, where the input documents are relatively structured, and there is no interactive behavior in such documents.

%Previous work~\cite{Goo18} showed that when jointly training two tasks, introducing a gate to explicitly share information(not only implicitly sharing information by using same LSTM parameters) between tasks can further improve the performance for both tasks.
Therefore, this work focuses on how to effectively model the interactive signals such as dialogue acts for better dialogue summarization, where we introduce a sentence-gated mechanism to jointly model the explicit relationships between dialogue acts and summaries.
To the best of our knowledge, there is no previous study with the similar idea, and we summarize our contributions as three-fold: 
\begin{compactitem}
\item The proposed model is the first attempt for dialogue summarization using dialogue acts as explicit interactive signals.
\item We benchmark the dataset for abstractive summarization in the meeting domain, where the summaries describe the high-level goals of meetings.
\item Our proposed model achieves the state-of-the-art performance in dialogue summarization and helps us analyze how much each utterance and its dialogue act affect the summaries.
%\item The gating results help us to know how much the topic can influence results of dialogue act prediction.
\end{compactitem}

\section{Dialogue Summarization Dataset}
\label{sec:data}

Considering that there is no abstractive summarization data in any conversational domain, this paper first builds a dataset in order to benchmark the experiments.
The AMI meeting corpus is a well-known meeting data with different annotations~\cite{mccowan2005ami}, which consists of 100 hours of meeting recordings.
The recordings use a range of signals synchronized to a common timeline, including close-talking and far-field microphones, individual and room-view video cameras, and output from a slide projector and an electronic whiteboard.
The meetings are recorded in English using three different rooms with different acoustic properties, and include mostly non-native speakers.
It contains a wide range of annotations, including dialogue acts, topic descriptions, named entities, hand gesture, and gaze direction.
In this work, we use the recording transcripts as the input to our model.
Because there is no summary annotation in the AMI data, the annotated topic descriptions are treated as summaries of the dialogues.
In AMI data, the annotations for dialogue acts and topic descriptions are not available for all utterances, so we extract a subset of the AMI corpus to construct the benchmark dialogue summarization dataset.
Figure~\ref{fig:example_data} is an example dialogue instance, where the summary describes the high-level goal of the meeting.

%\subsection{Preprocessing}
We use a sliding window size of 50 words to split a meeting into several dialogue samples, where we adjust the boundary to make sure no utterance would be cut in the middle.
%That is, if the word counts has reach 50, we will cut off the sample as soon as the speaker finished his/her sentence.
If the topic changes within the window, all topic descriptions are concatenated according to their appearing order.
In each resulting sample, there are around 50 to 100 words in an arbitrary number of sentences.
We extract 7,824 samples from 36 meeting recordings and then randomly separate them into three groups: 7,024 samples for training, 400 samples for development, and 400 samples for testing.
There are 15 dialogue act labels in the training set.
The detailed statistics are shown in Table~\ref{tab:dataset}.

\begin{table}
\centering
\begin{tabular}{|l|r|}
  \hline
   \bf AMI Corpus & \bf Statistics\\
  \hline\hline
  Vocabulary Size  & 8,886 \\
  \hline
  \#Dialogue Act & 15 \\
  \hline
  Min Summary Length & 1\\
  \hline
  Max Summary Length & 26\\
  \hline
  Training Set Size & 7,024  \\
  \hline
  Development Set Size & 400  \\
  \hline
  Testing Set Size & 400 \\
  \hline
\end{tabular}
\caption{Statistics of the AMI meeting corpus for dialogue summarization.}
\label{tab:dataset}
\end{table}

\begin{figure*}[t!]
\centering
\includegraphics[width=0.87\linewidth]{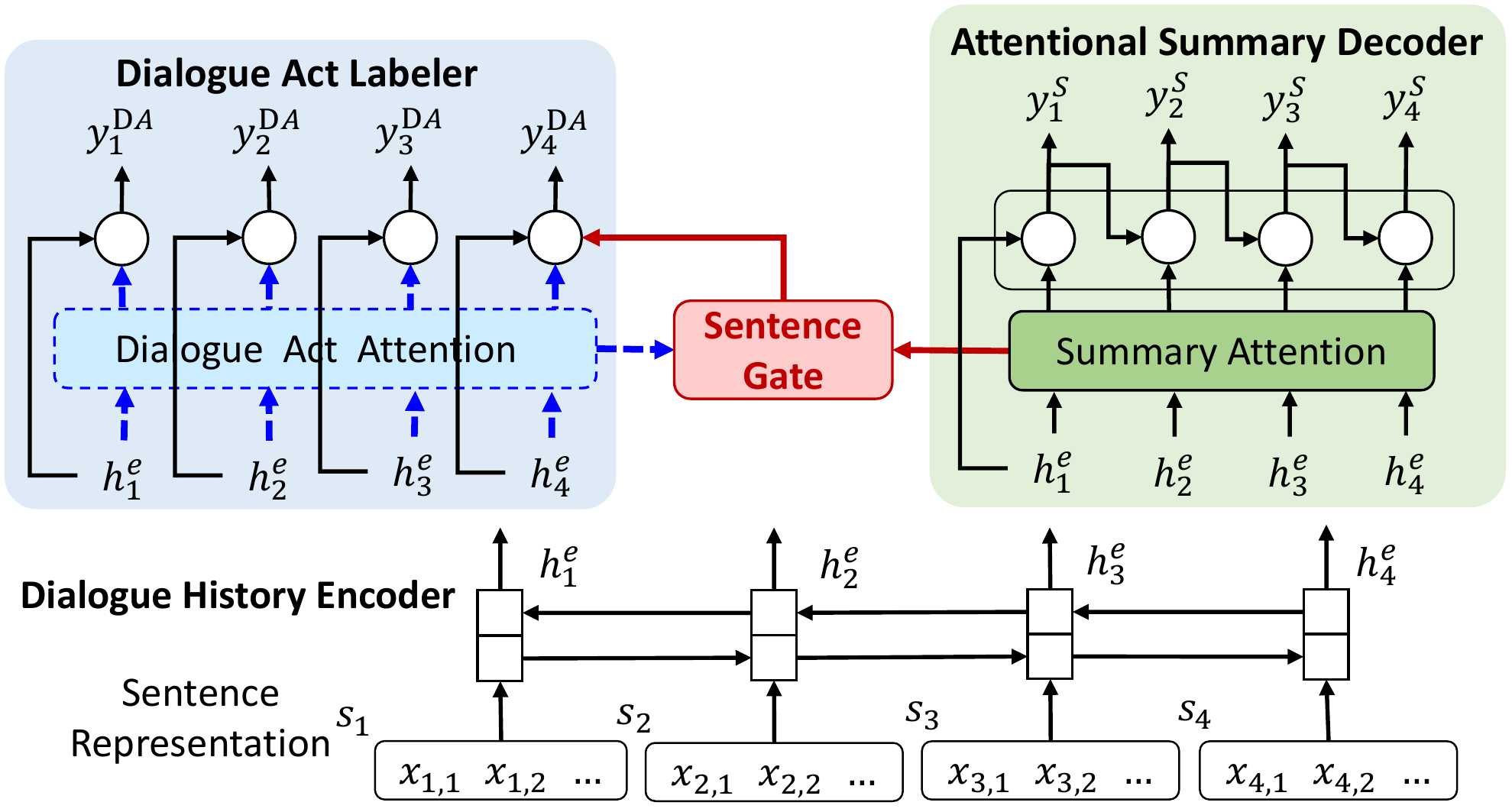}
\caption{The architecture of the proposed sentence-gated models.}
\label{fig:arch}
\end{figure*}

\section{Proposed Approach}
\label{sec:proposed}

This section first explains our attention-based RNN model and then introduces the proposed sentence gating mechanism for summarization jointly optimized with dialogue act recognition.
The model architecture is illustrated in Figure~\ref{fig:arch}, where there are several modules including 1) a \emph{dialogue history encoder}, 2) a \emph{dialogue act labeler}, 3) an \emph{attentional summary decoder}, and 4) a \emph{sentence gate}.
We detail each module below.

\subsection{Dialogue History Encoder}

Given a dialogue document, there is a sequence of utterances ${\bf s} = (s_1, \dots, s_K)$ as the input, where $K$ is the dialogue length.
An utterance is constituted by a word sequence ${\bf x} =(x_1,\dots,x_T)$, and the sentence embedding can be obtained by averaging all word embeddings in that sentence\footnote{The experiments using RNN-learned sentence embeddings are conducted, but the performance is similar to using the average of word embeddings. Considering the parameter size, all experiments use average vectors as sentence embeddings}.
The bidirectional long short-term memory (BLSTM) model~\cite{mesnil2015using} takes a sentence sequence ${\bf s} $ as the input, and then generates a forward hidden state $\overrightarrow{h_i^e}$ and a backward hidden state $\overleftarrow{h_i^e}$.
The final hidden state $h_i^e$ at the time step \textit{i} is the concatenation of $\overrightarrow{h_i^e}$ and $\overleftarrow{h_i^e}$, i.e. $h_i^e=[\overrightarrow{h_i^e},\overleftarrow{h_i^e}]$, which can be viewed as the encoded information for the given source document.

\subsection{Dialogue Act Labeler}
To leverage the dialogue act information, this module focuses on predicting dialogue acts for all utterances.
Specifically, $\textbf{s}$ is mapping to its corresponding dialogue act label ${\bf y} =(y^{DA}_1, \dots, y^{DA}_K)$.
For each hidden state $h_i$, we compute the dialogue act context vector $c_i^{DA}$ as the weighted sum of LSTM's hidden states, $h_1^e,...,h_T^e$, by the learned attention weights $\alpha^{DA}_{i, j}$:
\begin{equation}
c_i^{DA}=\sum_{j=1}^K{\alpha^{DA}_{i,j} \cdot h_j^e},
\end{equation}
where the dialogue act attention weights are computed as 
\begin{equation}
\alpha^{DA}_{i,j}=\frac{\exp(e_{i,j})}{\sum_{k=1}^K{\exp(e_{i,k})}}, 
\end{equation}
\begin{equation}
e_{i,k}=\sigma(W^{DA}_{he} \cdot h_k^e),
\end{equation}
where $\sigma$ is the \texttt{sigmoid} activation function, and $W^{DA}_{he}$ is the weight matrix of a feed-forward neural network.
Then all hidden states and dialogue act context vectors are optimized for dialogue act modeling by
\begin{equation}
\label{eq:da}
y^{DA}_i=\texttt{softmax}(W^{DA}_{hy}\cdot(h_i^e + c_i^{DA})),
\end{equation}
where $y^{DA}_i$ is the dialogue act label of the $i$-th sentence in the given dialogue, and $W^{DA}_{hy}$ is the weight matrix.
The dialogue act attention is shown as the blue component in Figure~\ref{fig:arch}.

\subsection{Attentional Summary Decoder}
Following the prior work~\cite{see2017get,cohan2018discourse}, we use an attentional decoder for generating the word sequence as the summary.
The summary context vector $c^S_i$ is computed as $c^{DA}$ similarly:
\begin{equation}
c^{S}_i=\sum_{j=1}^K{\alpha^{S}_{i,j} \cdot h_j^e}.
\end{equation}
The summary is generated by a unidirectional LSTM with the initial state set to be $h_K^e$, the last hidden state of the dialogue history encoder.
The unidirectional LSTM will output words until generating an end-of-string token or reaching the predefined maximum length.
The formulation is shown as:
\begin{equation}
\label{eq:sum_y}
y^S_i=\texttt{softmax}(W^S_{hy}\cdot (h_i^d + c^S_i)).
\end{equation}

\subsection{Sentence-Gated Mechanism}
\label{ssec:sent_gate}

\begin{figure}[t]
\centering
\includegraphics[width=.8\linewidth]{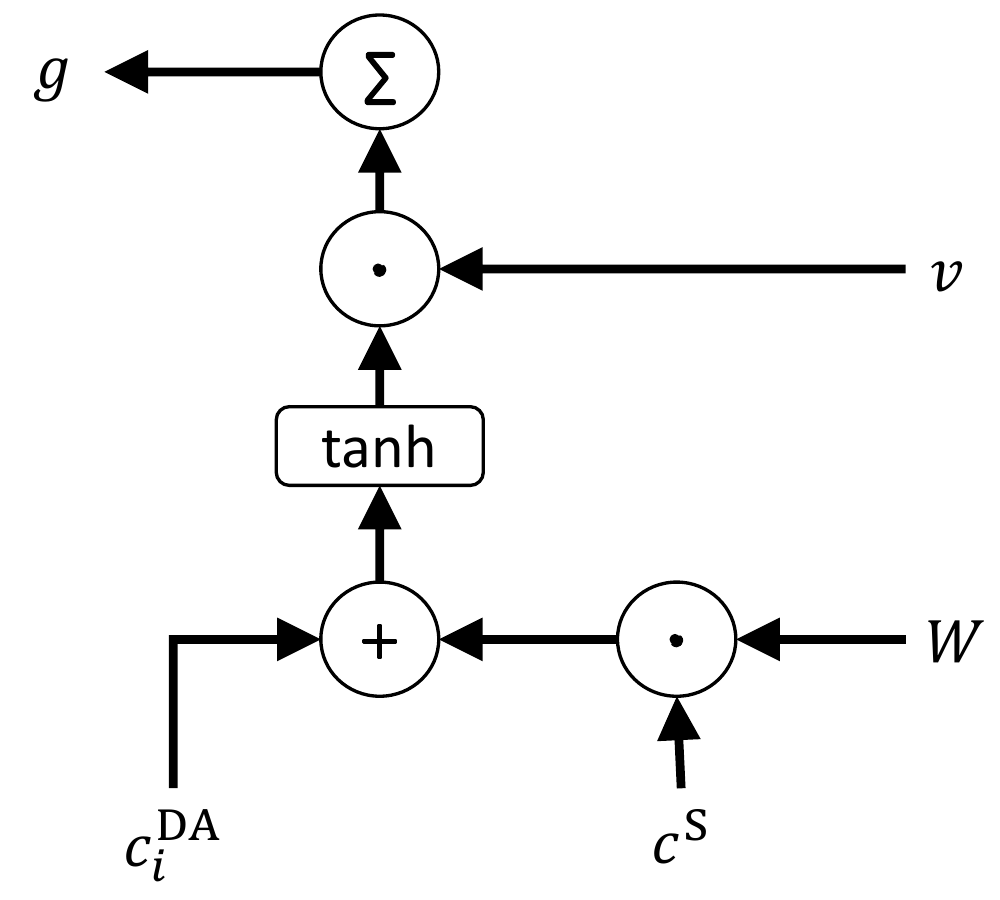}
\caption{Illustration of the sentence gate.}
\label{fig:gate}
\end{figure}

%This section introduces the proposed sentence-gated mechanism, which
%To model the explicit relationship between dialogue acts as interactive signal and the summary,
A gating mechanism is able to model the explicit relationship between two types of information~\cite{goo2018slot}.
The proposed sentence-gated model introduces an additional gate that leverages a summary context vector for modeling relationships between dialogue acts and summaries in order to improve the dialogue act labeler and the attentional summary decoder illustrated in Figure~\ref{fig:gate}.
The proposed model has two different types:
\begin{compactitem}
\item Full attention\\
The model considers the relations from dialogue acts and summaries using both \emph{dialogue act attention} and \emph{summary attention} shown as the blue and green blocks respectively in Figure~\ref{fig:arch}.
\item Summary attention\\
The model builds the gating mechanism using only \emph{summary attention}, where the parameter size is smaller than the full attention model.
\end{compactitem}

\subsubsection{Full Attention}
%(a) is one with both dialogue act attention and summary attention and (b) is another with only summary attention.
First, a dialogue act context vector $c_i^{DA}$ and an averaged summary context vector $c^S$ are combined to pass through a slot gate:
\begin{equation}
c^S=\frac{1}{K}\sum_{k=1}^K c^S_i,
\end{equation}
\begin{equation} \label{eq:gate} 
g=\sum{v\cdot \tanh(c_i^{DA} + W \cdot c^S)},
\end{equation}
where $v$ and $W$ are a trainable vector and a matrix respectively.
The summation is done over elements in one time step.
$g$ can be seen as a weighted feature of the joint context vector ($c_i^{DA}$ and $c^S$).
We use $g$ to weight between $h_i$ and $c_i^{DA}$ for deriving $y^{DA}_i$ and then replace (\ref{eq:da}) as below:
\begin{equation} \label{eq:da_y}
y^{DA}_i=\texttt{softmax}(W^{DA}_{hy}\cdot (h_i + c_i^{DA}\cdot g)).
\end{equation}
A larger $g$ indicates that the dialogue act context vector and the summary context vector pay attention to the similar part of the input sequence, which also infers that the correlation between the dialogue act and the summary is stronger and the context vector is more \emph{reliable} for contributing the prediction results.

\subsubsection{Summary Attention}
To deeply investigate the power of the sentence gate mechanism, we eliminate the dialogue act attention module in the architecture, so $c^{DA}_i$ is replaced with $h^e_i$.
%also propose a sentence-gated model with only summary attention in which 
Accordingly, (\ref{eq:gate}) and (\ref{eq:da_y}) are reformed as (\ref{eq:gate2}) and (\ref{eq:da_y2}) respectively,
%to allow only the summary attention to remain: 
\begin{equation}
\label{eq:gate2}
g=\sum{v\cdot\tanh(h^e_i+W\cdot c^S)}
\end{equation}
\begin{equation}
\label{eq:da_y2}
y^{DA}_i=\texttt{softmax}(W^{DA}_{hy}\cdot (h_i + h_i\cdot g))
\end{equation}
This version allows the dialogue acts and summaries to share the attention mechanism, so both information would be mutually improved in a more direct manner compared to the full attention version.

\subsection{Joint End-to-End Training}
To learn the summarization model optimized by the dialogue act information, we formulate a joint objective as
%obtain both dialogue act recognition and summarization jointly, the objective is formulated as
\begin{align}
 p(y^{DA}&, y^S\mid \textbf{s})\\
&=\prod_{k=1}^K{p(y^{S}_t\mid s_k)}\prod_{k=1}^K{p(y^{DA}_t\mid s_k)}\nonumber\\
&=\prod_{k=1}^K{p(y^{S}_t\mid \textbf{x}_k)}\prod_{k=1}^K{p(y^{DA}_t\mid \textbf{x}_k)},\nonumber
\end{align}
where $p(y^{DA}, y^S\mid \textbf{s})$ is the conditional probability of dialogue acts and the summary given the input dialogue.
Based on the joint objective, the proposed model that utilizes interactive signals for summarization can be trained in an end-to-end fashion.

\begin{table*}[!ht]
\centering
\begin{tabular}{|c|l||c||r||c|c|c|c|c|}
  \hline
  \multicolumn{2}{|c||}{\multirow{2}{*}{\bf Model}} & \bf \small Interactive & \multicolumn{1}{c||}{\multirow{2}{*}{\bf Size}} & \bf DA & \multicolumn{4}{c|}{\bf Summarization}\\
  \cline{5-9}
  \multicolumn{2}{|c||}{} & \bf \small Signal  && \bf Acc & \bf R-1 & \bf R-2 & \bf R-3 & \bf R-L\\
  \hline
  \hline
  \multicolumn{2}{|l||}{BLSTM Dialogue Act Labeler} & $\triangle$ & 3,864K & 64.16 & --- & --- & --- & --- \\
  \hline
%  \multicolumn{2}{|l||}{Attentional Seq2Seq~\cite{NallapatiXZ16}} & \xmark & 6,538K &--- & 42.48 & 32.34 & 27.45 & 42.48 \\
    \multicolumn{2}{|l||}{Attentional Seq2Seq~\cite{NallapatiXZ16}} & \xmark & 12,391K &--- & 34.74 & 25.15 & 21.35 & 34.70 \\
 % \multicolumn{2}{|l||}{Pointer-Generator Network~\cite{see2017get}} & \xmark &  7,319K & --- & 29.33 & 23.93 & 22.96 & 29.33 \\
  \multicolumn{2}{|l||}{Pointer-Generator Network~\cite{see2017get}} & \xmark  & 11,861K & --- & 31.21 & 26.35 & 25.22 & 31.21 \\
    \multicolumn{2}{|l||}{Discourse-Aware Hierarchical Seq2Seq~\cite{cohan2018discourse}} & $\triangle$ & 11,295K & --- & 66.82 & 37.74 & 27.71 & 47.84 \\
  \hline
  \multirow{2}{*}{Proposed} & Sentence-Gated (Full Attention) & \cmark & 12,363K & \bf 64.47$^\dag$ & 67.52$^\dag$ & 37.38 & 27.70 & 48.45$^\dag$ \\
 & Sentence-Gated (Summary)  & \cmark & 11,837K & 64.28 & \bf 68.34$^\dag$ & \bf 39.25$^\dag$ & \bf 29.05$^\dag$ & \bf 49.93$^\dag$ \\
  \hline
%  \hline
\end{tabular}
\caption{Performance on the AMI meeting data (\%). $^\dag$ indicates the significant improvement over all baselines with $p < 0.05$.} % and --- means the unavailable result.}
\label{tab:all_result}
\end{table*}

\section{Experiments}
\label{sec:experiment}

To evaluate the proposed model, we conduct experiments using the AMI meeting data introduced in Section~\ref{sec:data}.

\subsection{Setup}

In all experiments, the optimizer is \texttt{adam}, the reported numbers are averaged over 20 runs, and the maximum epoch is set to 30 with an early-stop strategy.
In our proposed model, the size of hidden vectors are set to 256, and the vector dimensions vary for the compared baselines such that all models have the similar size.
%for experiments shown in Table~\ref{tab:all_result} while the size of vectors vary for those shown in Table~\ref{tab:approximate_result} so that the number of parameters used in different models are approximately same.

For evaluation metrics, the dialogue act performance is measured by the accuracy (Acc), and the summary performance is measured by ROUGE-1 (R-1), ROUGE-2 (R-2), ROUGE-3 (R-3), and ROUGE-L (R-L) scores~\cite{lin2004rouge}.
We also validate the performance improvement with a statistical significance test for all experiments, where single-tailed t-test is performed to measure whether the results from the proposed model are significantly better than all baselines.
The dag symbols indicate the significant improvement with $p<0.05$.

\iffalse
\begin{table*}[!ht]
\centering
\begin{tabular}{|c|l||c|c||r||c|c|c|c|c|}
  \hline
  \multicolumn{2}{|c||}{\multirow{2}{*}{\bf Model}} & \bf \small Interact & \bf \small Pointer & \multicolumn{1}{c||}{\multirow{2}{*}{\bf Size}} & \bf DA & \multicolumn{4}{c|}{\bf Summarization}\\
  \cline{6-10}
  \multicolumn{2}{|c||}{} & \bf \small Signal & \bf \small Copy && \bf Acc & \bf R-1 & \bf R-2 & \bf R-3 & \bf R-L\\
  \hline
  \hline
  \multicolumn{2}{|l||}{Dialogue Act BLSTM} &&& 10,873K & 63.98 & --- & --- & --- & --- \\
  \hline
  \multicolumn{2}{|l||}{Attentional Seq2Seq~\cite{NallapatiXZ16}} & \xmark & \xmark & 12,391K &--- & 34.74 & 25.15 & 21.35 & 34.70 \\
  \multicolumn{2}{|l||}{Discourse-Aware Hierarchical Seq2Seq~\cite{cohan2018discourse}} & $\triangle$ & \xmark & 11,295K & --- & 66.82 & 37.74 & 27.71 & 47.84 \\
  \multicolumn{2}{|l||}{Pointer-Generator Network~\cite{see2017get}} & \xmark & \cmark & 11,861K & --- & 31.21 & 26.35 & 25.22 & 31.21 \\
  \hline
  \multirow{2}{*}{Proposed} & Sentence-Gated (Full Attention) & \cmark & \xmark & 12,363K & \bf 64.47$^\dag$ & 67.52$^\dag$ & 37.38 & 27.70 & 48.45$^\dag$ \\
 & Sentence-Gated (Summary)  & \cmark &\xmark & 11,837K & 64.28 & \bf 68.34$^\dag$ & \bf 39.25$^\dag$ & \bf 29.05$^\dag$ & \bf 49.93$^\dag$ \\
  \hline
%  \hline
\end{tabular}
\caption{Performance on AMI meeting data (\%) with approximate parameter size. $^\dag$ indicates the significant improvement over all baselines ($p < 0.05$) and --- means the result is unavailable.}
\label{tab:approximate_result}
\end{table*}
\fi

\begin{figure*}[t]
\centering
\begin{tabular}{|p{0.83\linewidth}|c|}
\hline
\multicolumn{1}{|c|}{\bf Testing Dialogue Example 1} & \bf Dialogue Act\\
\hline\hline
A: okay . & Assess \\
B: okay that's fine , that's good . & Assess \\
C: \cellcolor{blue!40}okay , let's start from the beginning & \cellcolor{blue!40}Offer \\
C: \cellcolor{blue!50}so i'm going to speak about technical functions design & \cellcolor{blue!50}Inform \\
C: \cellcolor{blue!50}un just like some some first issues that came up . & \cellcolor{blue!50}Inform \\
B: um 'kay , & Stall \\
C: \cellcolor{blue!50}so the method i was um adopting  at this point , it's not um for the for the whole um period of the um all the project but it's just at th at this very moment . & \cellcolor{blue!50}Inform \\
B: um & Stall \\
C: \cellcolor{blue!50}uh my method was um to look at um other um remote controls , & \cellcolor{blue!50}Inform \\
C: \cellcolor{blue!60}uh so mostly just by searching on the web & \cellcolor{blue!60}Inform \\
C: \cellcolor{blue!50}and to see what um functionality they used . & \cellcolor{blue!50}Inform \\
C: \cellcolor{blue!30}and then um after having got this inspiration and having compared what i found on the web um just to think about what the de what the user really needs and what um what the use might desire as additional uh functionalities . & \multirow{3}{*}{\cellcolor{blue!30} Inform} \\
\hline
\multicolumn{2}{|l|}{Generated summary: industrial designer \underline{presentation} issues of participants} \\
\multicolumn{2}{|l|}{Reference summary: industrial designer presentation interface specialist presentation} \\
\hline
\hline
\multicolumn{1}{|c|}{\bf Testing Dialogue Example 2} & \bf Dialogue Act\\
\hline\hline
A: okay , so & Stall \\
B: hmm , okay . & Backchannel \\
A: yeah well uh & Stall \\
A: \cellcolor{blue!10}ipod is trendy . & \cellcolor{blue!10}Inform \\
A: \cellcolor{blue!40}and it is well curved square . & \cellcolor{blue!40}Inform \\
C: yeah . & Backchannel \\
A: \cellcolor{blue!20}square . like . & \cellcolor{blue!20}Inform \\
B: \cellcolor{blue!10}yeah , but mm is uh has round corners i think . & \cellcolor{blue!10}Assess \\
A: so & Stall \\
D: \cellcolor{blue!70}we shouldn't have too square corners and that kind of thing . & \cellcolor{blue!70}Inform \\
\hline
\multicolumn{2}{|l|}{Generated summary: \underline{look} and usability} \\
\multicolumn{2}{|l|}{Reference summary: look and usability} \\
\hline
\end{tabular}
\caption{Visualization of summary attention vectors. The darker color indicates higher attention wights. The underlined word is the target word for illustrating the attention.}
\label{fig:example_attn}
\end{figure*}

\subsection{Baselines}
Considering that there is no previous work for joint dialogue act modeling and summarization, the compared baselines are either for dialogue act classification or text summarization, including a bidirectional LSTM for dialogue act labeler, an attentional seq2seq summarization model~\cite{NallapatiXZ16}, a pointer-generator network~\cite{see2017get}, and a discourse-aware hierarchical attentional seq2seq~\cite{cohan2018discourse}.
Please note that the BLSTM dialogue act labeler baseline is the same as our proposed model without the summarization component.
The pointer-generator network extends the attentional seq2seq by adding a joint pointer network to enable the copy mechanism,
For the discourse-aware model, we only use the concept about the hierarchy introduced by Cohan et al.~\cite{cohan2018discourse} but do not include its pointer network part.
The reason will be latter explained in Section~\ref{sec:result}.
%, , and the bidirectional LSTM for dialogue act is our proposed model without the summarization part.
Among all baselines, only the discourse-aware model implicitly utilizes the interactive signal, while our model explicitly optimizes the summary together with dialogue acts.

\subsection{Results}
\label{sec:result}

The experimental results are shown in Table~\ref{tab:all_result}, where the models have similar size of parameters.
Among all summarization baselines, the discourse-aware hierarchical seq2seq model achieves better performance than other two baselines, indicating the importance of discourse/interaction cues for dialogue summarization.
%We think that knowing sentence boundary is a crucial point for summary model, so hierarchy attention model is much better than the other two summary baselines.
Comparing between attentional seq2seq and the pointer-generator network, the difference is not obvious, because the high-level descriptions as summaries do not overlap between the input dialogues and the corresponding summaries (1.2\% of the overlapping rate for AMI meeting data).
%word overlap: train 1.20% test 1.15% valid 1.20%
Therefore, due to the low overlapping rate, the pointer-generator network performs the worst, because the pointer network and coverage loss parts introduce noises.
This is the reason that other baselines and our proposed model do not contain the copy mechanism and coverage loss in the experiments.
The finding suggests that the dialogue summarization focuses more on the \emph{interaction goal} instead of the \emph{mentioned content}.
%Also in this meeting scenario, the summary can be viewed as a high level description of conversation, there is few things pointer network and coverage loss can do which leads Pointer-Generator network to have the worst performance.

Table~\ref{tab:all_result} shows that the proposed sentence-gated mechanism with summary attention significantly outperforms all baselines, where almost all measurements obtain the significant improvement, demonstrating that interactive signal provides useful cues for dialogue summarization, and the proposed sentence-gated mechanism effectively models the relationships between them.
The proposed model with full attention performs slightly worse than the one with summary attention only.
The probable reason is that the dialogue act attention may not be necessary for predicting the dialogue acts of a single utterance; that is, dialogue acts are often decided only based on the individual utterance, so adding attention on its contextual utterances may not bring much benefit for modeling such interactive behaviors.
Moreover, the proposed model reduces the model size by 12\% compared to the best baseline combination (BLSTM for dialogue act prediction + discourse-aware hierarchical attention seq2seq for summarization) and demonstrates the better model capacity.
%explain why full attention does not work
%Because AMI meeting corpus only uses hundreds of words to generate topic summary, we encounter some over-fitting issue when training the model.
%Figure~\ref{fig:word_dist} plots the word frequency distribution of summary.
%Proposed full-attention model performs worse than the summary-attention-only model may due to this reason.

\subsection{Attention Analysis}

To further analyze the attention learned in the model, we visualize the utterance attention weights when generating summaries in Figure~\ref{fig:example_attn}.
Figures are colored with different levels of the summary attention, where the darker one has a larger attention value as its importance when generating the target word, and vice versa.
It is obvious that the proposed model successfully captures which ones are the key sentences in the dialogues.
It may be credited to the proposed sentence gate that learns the dialogue acts conditioned on its summary in order to provide the helpful signal for global optimization of the joint model.
In addition, it can be found that the ``Inform'' dialogue act usually guides the model to pay more attention to it, which aligns well with our intuition.
In sum, for dialogue summarization, the experiments show that modeling dialogue acts and summary relations controlled by the novel sentence-gated mechanism can effectively improve abstractive summarization performance in terms of ROUGE scores due to the joint optimization with dialogue act modeling.

%\begin{figure}[ht]
%\centering
%\includegraphics[width=\linewidth]{summary_word_dist}
%\caption{Summary Word Distribution.}
%\label{fig:word_dist}
%\end{figure}

\section{Conclusion}
\label{sec:conclude}

This paper focuses on abstractive dialogue summarization by modeling interactive behaviors, where the proposed model uses a novel sentence-gate that allows the dialogue act signal can be conditioned on the learned summarization result, in order to achieve better performance for both tasks.
This paper benchmarks the experiments using a meeting dataset, and the experiments show that the proposed approach outperforms all state-of-the-art models, demonstrating the importance of interactive cues in dialogue summarization.

\section{Acknowledgements}
We thank the anonymous reviewers for their insightful feedback on this work.
The authors are financially supported by Ministry of Science and Technology (MOST) in Taiwan and MediaTek Inc.

%The idea of the gated mechanism can be extended to different tasks.
%Also, the sentence-gated model is more useful for understanding a dialogue, because the dialogue act is now conditioned on conversation topics, which is easier to model, and this paper provides the guidance of model design for future dialogue system.

\newpage
\bibliographystyle{IEEEbib}
\bibliography{refs}

\begin{thebibliography}{10}

\bibitem{Kupiec1995}
Julian Kupiec, Jan Pedersen, and Francine Chen,
\newblock ``A trainable document summarizer,''
\newblock in {\em Proceedings of SIGIR}, 1995, pp. 68--73.

\bibitem{chen2011spoken}
Yun-Nung Chen, Yu~Huang, Ching-Feng Yeh, and Lin-Shan Lee,
\newblock ``Spoken lecture summarization by random walk over a graph
  constructed with automatically extracted key terms,''
\newblock in {\em Proceedings of INTERSPEECH}, 2011, pp. 933--936.

\bibitem{saggion2012}
Horacio Saggion and Thierry Poibeau,
\newblock ``{Automatic Text Summarization: Past, Present and Future},''
\newblock in {\em {Multi-source, Multilingual Information Extraction and
  Summarization}}, R.~Yangarber T.~Poibeau; H. Saggion. J.~Piskorski, Ed.,
  Theory and Applications of Natural Language Processing, pp. 3--13.
  {Springer}, 2012.

\bibitem{liu2015towards}
Fei Liu, Jeffrey Flanigan, Sam Thomson, Norman Sadeh, and Noah~A Smith,
\newblock ``Toward abstractive summarization using semantic representations,''
\newblock in {\em Proceedings of NAACL-HLT}, pp. 1077--1086.

\bibitem{chopra2016abstractive}
Sumit Chopra, Michael Auli, and Alexander~M Rush,
\newblock ``Abstractive sentence summarization with attentive recurrent neural
  networks,''
\newblock in {\em Proceedings of NAACL-HLT}, 2016, pp. 93--98.

\bibitem{NallapatiXZ16}
Ramesh Nallapati, Bing Xiang, and Bowen Zhou,
\newblock ``Sequence-to-sequence rnns for text summarization,''
\newblock {\em CoRR}, 2016.

\bibitem{rush2015neural}
Alexander~M Rush, Sumit Chopra, and Jason Weston,
\newblock ``A neural attention model for abstractive sentence summarization,''
\newblock in {\em Proceedings of EMNLP}, 2015, pp. 379--389.

\bibitem{gu2016}
Jiatao Gu, Zhengdong Lu, Hang Li, and Victor~O.K. Li,
\newblock ``Incorporating copying mechanism in sequence-to-sequence learning,''
\newblock in {\em Proceedings of ACL}, Berlin, Germany, 2016, pp. 1631--1640.

\bibitem{miao2016}
Yishu Miao and Phil Blunsom,
\newblock ``Language as a latent variable: Discrete generative models for
  sentence compression,''
\newblock in {\em Proceedings of EMNLP}, 2016, pp. 319--328.

\bibitem{see2017get}
Abigail See, Peter~J Liu, and Christopher~D Manning,
\newblock ``Get to the point: Summarization with pointer-generator networks,''
\newblock in {\em Proceedings of ACL}, 2017, vol.~1, pp. 1073--1083.

\bibitem{hsu2018unified}
Wan-Ting Hsu, Chieh-Kai Lin, Ming-Ying Lee, Kerui Min, Jing Tang, and Min Sun,
\newblock ``A unified model for extractive and abstractive summarization using
  inconsistency loss,''
\newblock in {\em Proceedings of ACL}, 2018, pp. 1--10.

\bibitem{lee2014spoken}
Hung-yi Lee, Sz-Rung Shiang, Ching-feng Yeh, Yun-Nung Chen, Yu~Huang, Sheng-Yi
  Kong, and Lin-shan Lee,
\newblock ``Spoken knowledge organization by semantic structuring and a
  prototype course lecture system for personalized learning,''
\newblock {\em IEEE/ACM Transactions on Audio, Speech and Language Processing},
  vol. 22, no. 5, pp. 883--898, 2014.

\bibitem{gehrmann2018bottom}
Sebastian Gehrmann, Yuntian Deng, and Alexander~M Rush,
\newblock ``Bottom-up abstractive summarization,''
\newblock in {\em Proceedings of EMNLP}, 2018.

\bibitem{maskey2005comparing}
Sameer Maskey and Julia Hirschberg,
\newblock ``Comparing lexical, acoustic/prosodic, structural and discourse
  features for speech summarization,''
\newblock in {\em Proceedings of EUROSPEECH}, 2005.

\bibitem{harwath2012topic}
David Harwath and Timothy~J Hazen,
\newblock ``Topic identification based extrinsic evaluation of summarization
  techniques applied to conversational speech,''
\newblock in {\em Proceedings of ICASSP}, 2012, pp. 5073--5076.

\bibitem{riedhammer2010long}
Korbinian Riedhammer, Benoit Favre, and Dilek Hakkani-T{\"u}r,
\newblock ``Long story short--global unsupervised models for keyphrase based
  meeting summarization,''
\newblock {\em Speech Communication}, vol. 52, no. 10, pp. 801--815, 2010.

\bibitem{chen2011automatic}
Yun-Nung Chen,
\newblock ``Automatic key term extraction and summarization from spoken course
  lectures,''
\newblock M.S. thesis, National Taiwan University, 6 2011.

\bibitem{chen2012two}
Yun-Nung Chen and Florian Metze,
\newblock ``Two-layer mutually reinforced random walk for improved multi-party
  meeting summarization,''
\newblock in {\em Proceedings of SLT}, 2012, pp. 461--466.

\bibitem{chen2013multi}
Yun-Nung Chen and Florian Metze,
\newblock ``Multi-layer mutually reinforced random walk with hidden parameters
  for improved multi-party meeting summarization,''
\newblock in {\em Proceedings of INTERSPEECH}, 2013, pp. 485--489.

\bibitem{chen2012intra}
Yun-Nung Chen and Florian Metze,
\newblock ``Intra-speaker topic modeling for improved multi-party meeting
  summarization with integrated random walk,''
\newblock in {\em Proceedings of NAACL-HLT}, 2012, pp. 377--381.

\bibitem{gambhir2017recent}
Mahak Gambhir and Vishal Gupta,
\newblock ``Recent automatic text summarization techniques: a survey,''
\newblock {\em Artificial Intelligence Review}, vol. 47, no. 1, pp. 1--66,
  2017.

\bibitem{mccowan2005ami}
Iain McCowan, Jean Carletta, W~Kraaij, S~Ashby, S~Bourban, M~Flynn,
  M~Guillemot, T~Hain, J~Kadlec, V~Karaiskos, et~al.,
\newblock ``The ami meeting corpus,''
\newblock in {\em Proceedings of the 5th International Conference on Methods
  and Techniques in Behavioral Research}, 2005, vol.~88, p. 100.

\bibitem{Bunt94}
Harry Bunt,
\newblock ``Context and dialogue control,''
\newblock {\em THINK Quarterly}, vol. 3, 1994.

\bibitem{Samuel1998}
Ken Samuel, Sandra Carberry, and K.~Vijay-Shanker,
\newblock ``Dialogue act tagging with transformation-based learning,''
\newblock in {\em Proceedings of COLING}, 1998.

\bibitem{wright1999}
Helen Wright, Massimo Poesio, and Stephen Isard,
\newblock ``Using high level dialogue information for dialogue act recognition
  using prosodic features,''
\newblock in {\em ESCA Tutorial and Research Workshop (ETRW) on Dialogue and
  Prosody}, 1999.

\bibitem{Stolcke00}
Andreas Stolcke, Noah Coccaro, Rebecca Bates, Paul Taylor, Carol Van
  Ess-Dykema, Klaus Ries, Elizabeth Shriberg, Daniel Jurafsky, Rachel Martin,
  and Marie Meteer,
\newblock ``Dialogue act modeling for automatic tagging and recognition of
  conversational speech,''
\newblock {\em Computational Linguistics}, vol. 26, no. 3, pp. 339--373, Sept.
  2000.

\bibitem{Kluwer2010}
Tina Kl\"{u}wer, Hans Uszkoreit, and Feiyu Xu,
\newblock ``Using syntactic and semantic based relations for dialogue act
  recognition,''
\newblock in {\em Proceedings of COLING}, 2010, pp. 570--578.

\bibitem{tran2017preserving}
Quan~Hung Tran, Ingrid Zukerman, and Gholamreza Haffari,
\newblock ``Preserving distributional information in dialogue act
  classification,''
\newblock in {\em Proceedings of EMNLP}, 2017, pp. 2151--2156.

\bibitem{tavafi2013}
Maryam Tavafi, Yashar Mehdad, Shafiq Joty, Giuseppe Carenini, and Raymond Ng,
\newblock ``Dialogue act recognition in synchronous and asynchronous
  conversations,''
\newblock in {\em Proceedings of SIGDIAL}, 2013, pp. 117--121.

\bibitem{Keizer2002}
Simon Keizer, Rieks op~den Akker, and Anton Nijholt,
\newblock ``Dialogue act recognition with bayesian networks for dutch
  dialogues,''
\newblock in {\em Proceedings of SIGDIAL}, 2002, pp. 88--94.

\bibitem{Ang05}
J.~Ang, Yang Liu, and E.~Shriberg,
\newblock ``Automatic dialog act segmentation and classification in multiparty
  meetings,''
\newblock in {\em Proceedings of ICASSP}, 2005, pp. 1061--1064.

\bibitem{chen2013empirical}
Yun-Nung Chen, William~Yang Wang, and Alexander~I Rudnicky,
\newblock ``An empirical investigation of sparse log-linear models for improved
  dialogue act classification,''
\newblock in {\em Proceedings of ICASSP}. IEEE, 2013, pp. 8317--8321.

\bibitem{ji16}
Yangfeng Ji, Gholamreza Haffari, and Jacob Eisenstein,
\newblock ``A latent variable recurrent neural network for discourse-driven
  language models,''
\newblock in {\em Proceedings of NAACL-HLT}, 2016, pp. 332--342.

\bibitem{Khanpour16}
Hamed Khanpour, Nishitha Guntakandla, and Rodney~D. Nielsen,
\newblock ``Dialogue act classification in domain-independent conversations
  using a deep recurrent neural network,''
\newblock in {\em Proceedings of COLING}, 2016.

\bibitem{lee2016}
Ji~Young Lee and Franck Dernoncourt,
\newblock ``Sequential short-text classification with recurrent and
  convolutional neural networks,''
\newblock in {\em Proceedings of NAACL-HLT}, 2016, pp. 515--520.

\bibitem{Kalchbrenner13}
Nal Kalchbrenner and Phil Blunsom,
\newblock ``Recurrent convolutional neural networks for discourse
  compositionality,''
\newblock in {\em Proceedings of the Workshop on Continuous Vector Space Models
  and their Compositionality}, 2013, pp. 119--126.

\bibitem{cohan2018discourse}
Arman Cohan, Franck Dernoncourt, Doo~Soon Kim, Trung Bui, Seokhwan Kim, Walter
  Chang, and Nazli Goharian,
\newblock ``A discourse-aware attention model for abstractive summarization of
  long documents,''
\newblock in {\em Proceedings of NAACL-HLT}, 2018, vol.~2, pp. 615--621.

\bibitem{mesnil2015using}
Gr{\'e}goire Mesnil, Yann Dauphin, Kaisheng Yao, Yoshua Bengio, Li~Deng, Dilek
  Hakkani-Tur, Xiaodong He, Larry Heck, Gokhan Tur, Dong Yu, et~al.,
\newblock ``Using recurrent neural networks for slot filling in spoken language
  understanding,''
\newblock {\em IEEE/ACM Transactions on Audio, Speech and Language Processing
  (TASLP)}, vol. 23, no. 3, pp. 530--539, 2015.

\bibitem{goo2018slot}
Chih-Wen Goo, Guang Gao, Yun-Kai Hsu, Chih-Li Huo, Tsung-Chieh Chen, Keng-Wei
  Hsu, and Yun-Nung Chen,
\newblock ``Slot-gated modeling for joint slot filling and intent prediction,''
\newblock in {\em Proceedings of NAACL-HLT}, 2018, pp. 753--757.

\bibitem{lin2004rouge}
Chin-Yew Lin,
\newblock ``{ROUGE}: A package for automatic evaluation of summaries,''
\newblock {\em Text Summarization Branches Out}, 2004.

\end{thebibliography}

\end{document}